\documentclass[lettersize,journal]{IEEEtran}
\usepackage{amsmath,amsfonts}
\usepackage{algorithmic}
\usepackage{array}
\usepackage[caption=false,font=normalsize,labelfont=sf,textfont=sf]{subfig}
\usepackage{textcomp}
\usepackage{stfloats}
\usepackage{url}
\usepackage{verbatim}
\usepackage{graphicx}
\usepackage{xcolor}
\usepackage{cite}
\usepackage{hyperref}
\usepackage{threeparttable}
\usepackage{multirow}
\usepackage{booktabs}
\def\BibTeX{{\rm B\kern-.05em{\sc i\kern-.025em b}\kern-.08em
    T\kern-.1667em\lower.7ex\hbox{E}\kern-.125emX}}
\usepackage{balance}
\begin{document}
\title{In-Context Learning World Models for Robot Dual-Arm Manipulation of Soft Garment with Partial Observations}

\author{Hengyi Sim, Quentin Rouxel, Dennis Hong,~\IEEEmembership{Member,~IEEE},  \\Darwin Caldwell,~\IEEEmembership{Fellow,~IEEE} and Fei Chen, \IEEEmembership{Senior Member, IEEE}
\vspace{-2em}
\thanks{This work is supported in part by project \#MMT-p6-26 of the Shun Hing Institute of Advanced Engineering, The Chinese University of Hong Kong, in part by the Strategic Partnership Award for Research Collaboration, The Chinese University of Hong Kong, in part by the Research Grants Council of the Government of the Hong Kong SAR via the Grant 14213324, C7100-22GF and in part by the InnoHK of the Government of the Hong Kong SAR via the Hong Kong Centre for Logistics Robotics. $\textit{(Corresponding author: Fei Chen.)}$}
	\thanks{Hengyi Sim, Quentin Rouxel and Fei Chen are with the Collaborative and Versatile Robots (CLOVER) Laboratory, Department of Mechanical and Automation Engineering, T-Stone Robotics Institute, The Chinese University of Hong Kong, Hong Kong. Fei Chen is also affiliated with the Shun Hing Institute of Advanced Engineering, The Chinese University of Hong Kong (e-mail: {\tt\small  hysim@mae.cuhk.edu.hk, quentinrouxel@cuhk.edu.hk, f.chen@ieee.org}).}
    \thanks{Dennis Hong is with the Department of Mechanical and Aerospace Engineering, University of California, Los Angeles, USA (e-mail: {\tt\small dennishong@g.ucla.edu}).}
    \thanks{Darwin Caldwell is with the Department of Advanced Robotics, Istituto Italiano di Tecnologia, Genoa, Italy (e-mail: {\tt\small darwin.caldwell@iit.it}).}
}


\maketitle

\begin{abstract}
Robust manipulation of soft objects, such as garments, is a critical skill for service and household robots to adopt, but remains challenging due to its high-dimensional state space, and material-dependent deformations. Humans address this challenge by using mental world models that "imagine" the shape and material properties from partial observations, which enables them to predict how actions affect future states. As such, embedding such world models in robots allows for better generalization in manipulating a wide variety of garments in the real world. However, learned world models typically require large amounts of real world data that is strenuous to collect, and a mechanism to infer latent properties that govern dynamics. We present a Soft-Object World Model that performs generative full-state estimation and action-conditioned dynamics prediction using a transformer backbone trained on large-scale simulated soft object interactions, with measures to mitigate the sim-to-real gap. By learning to infer latent shape and material cues through in-context learning over a variable-length partial point-cloud history, the model generalizes across robot embodiments, material parameters, garment types and shapes, and manipulation coordination modes without weight updates. The world model is then integrated with a Goal-Conditioned action planner for motion synthesis. We validate the approach in simulation and real world for bimanual garment folding, reporting improvements in reconstruction fidelity, prediction accuracy, and task success. Project Website: \url{https://1am5hy.github.io/ICL_soft_object_world_model/}
\end{abstract}

\vspace{-1em}

\begin{figure}[!t]
	\centerline{\includegraphics[width=1\linewidth]{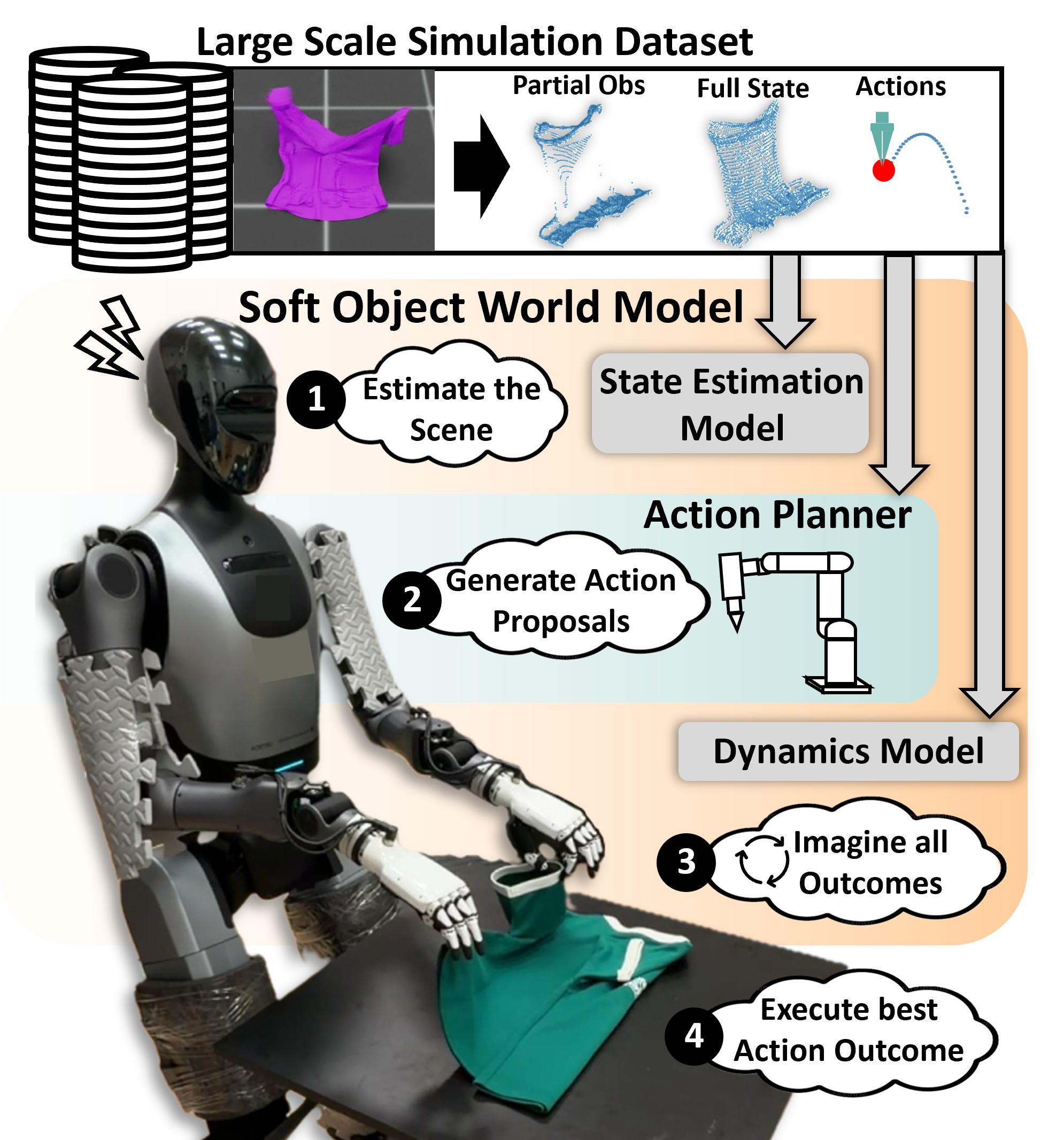}} 
     \vspace{-.75em}
	\caption{ This work leverages a large scale simulation dataset of soft-object point-cloud interactions  to perform generative full-state estimation and action-conditioned dynamics prediction, with In-Context Learning for online material identification. This work is evaluated on bimanual garment folding, across different robot embodiments, including a humanoid robot.}
	\label{front_page}
    \vspace{-1.75em}
\end{figure}

\section{Introduction}
Robotic manipulation of soft objects is fundamental to a wide range of domestic, service, healthcare, and industrial tasks, including assistive dressing \cite{kotsovolis2024garment} and laundry folding \cite{hoque2020visuospatial}, as seen in Fig \ref{front_page}. Despite its practical significance, robotic manipulation research has predominantly focused on rigid objects \cite{bohg2013data}, as soft-object manipulation remains substantially more challenging \cite{zhu2022challenges}. Soft objects exhibits high-dimensional state spaces and complex nonlinear deformations, while their configurations are often only partially observable because of self-occlusion \cite{delehelle2024garfield}. Moreover, their dynamics are highly sensitive to variations in geometry, material properties, and contact conditions. Humans address such challenges by forming internal predictive models acquired from a lifetime of data and experience. Such learned mental models can estimate hidden states, encode implicit knowledge of object geometry and material behavior, to subsequently predict the effects of actions on the current soft object state. This capacity for learning predictive representations from diverse experience motivates the use of data-driven models in robotics, as recent data-driven models have shown that large-scale predictive learning can extract transferable structure from diverse data  \cite{brown2020language,radford2021learning, young2018recent, yan2024generative, choong2023jack}. 

In robotics, learned world models have similarly proven effective for inferring latent state, forecasting future observations, and planning by simulating candidate trajectories within a learned dynamical manifold \cite{li2025robotic}. Extending this paradigm to soft‑object manipulation is therefore a natural step. Garments and other soft objects exhibits strong nonlinear, material‑dependent dynamics that are hard to capture analytically but are well suited to learned predictive models that support dynamics reasoning. However, robotics models lacks the large and varied real-world interaction datasets available in language and vision. Collecting real manipulation data is expensive, labor-intensive, and difficult to scale, particularly for soft objects that require repeated resetting and exhibit substantial variation in geometry and material behavior.

The first major challenge in building soft object world models is the high DOFs of soft object state space, where there exists many valid state configurations that occupy a complex manifold. In this space, small changes in contact point, geometry, or material parameters can cause different valid states with similar partial observations, further complicating state estimation. Partial observation from self-occlusion, folds, and camera poses routinely occludes task-critical regions of the soft object, forcing the world model to infer plausible unseen geometry. Selecting an effective state representation is therefore essential, as it must retain sufficient geometric detail for accurate prediction while remaining robust to sim‑to‑real mismatch. High-dimensional point-cloud representations are attractive because they preserve fine-grained geometry, map naturally to depth sensors, and empirically transfer well from physics simulators to real sensors when paired with appropriate domain randomization and generative conditioning \cite{tian2025diffusion}. However, the data requirements for learning such expressive models are large and collecting extensive real-world demonstrations is expensive, thus motivating large-scale synthetic corpus generation in simulators like Isaac Lab \cite{mittal2025isaac}. 

The second major challenge is dynamics prediction with rapid adaptation to material properties during inference. Predicting the dynamics of soft objects is challenging because deformations are strongly coupled to unknown material properties, grasp points, and the current unseen shape, where the same action can produce different outcomes \cite{11236994}. In-context learning (ICL) provides a mechanism to reduce this ambiguity at test time. By conditioning the model on a short set of unrelated prior demonstrations, ICL infers material and shape priors without gradient-based fine-tuning \cite{vosylius2024instant}. These demonstrations can also include diverse exaggerated  dynamics, thus improving the model's ability to estimate dynamic properties, enabling rapid adaptation and more reliable forward rollouts. Accurate, context-conditioned world models can then be plugged directly into planning and control loops, serving as forward models, as action proposal evaluators or model-predictive controllers, therefore enabling robots to reason about long-horizon outcomes and select manipulation strategies that are robust across diverse garments.

In this study, we aim to introduce and experimentally evaluate a Transformer-based soft object world model with test-time in-context adaptation for robotic manipulation. The framework reconstructs full object states from partial observations and predicts future deformations conditioned on planned actions and past estimated demonstration states. An in-context conditioning mechanism uses short sequences of estimated states to infer object material properties at inference without fine-tuning. To address data limitations and improve sim-to-real transfer, we train on a large-scale and diverse corpus of point-cloud-based soft object interactions, generated with Isaac Lab with randomized material parameters. We evaluate full-state reconstruction, multi-step action-conditioned dynamics prediction, and adaptation to variations in object geometry and material properties in simulation and real-world bimanual garment-manipulation experiments. Finally, we demonstrate sim-to-real transfer across multiple garment categories and material variants without test-time parameter updates.

\section{Related Work}
\subsection{Soft Object Manipulation}
Manipulating soft objects such as garments remains a long-standing challenge in robotics, due to their high-dimensional state space and complex, nonlinear dynamics. Some works reduce the control problem by manually planning the manipulation trajectory and restricting learning to grasp‑point selection \cite{ha2022flingbot}, an approach exemplified by methods that also predict manipulation primitive types alongside grasp selection \cite{zhou2025dual, avigal2022speedfolding}. Unfortunately, these methods and some others that rely on visuomotor servoing \cite{osawa2007unfolding, maitin2010cloth}, are constrained to the pre-defined skill trajectory. Data‑driven learning methods offer an alternative by enabling end‑to‑end mapping from unstructured perceptual inputs to control actions. 

Such methods like model-free reinforcement learning (RL) \cite{matas2018sim, jangir2020dynamic} and imitation learning (IL) \cite{chi2025diffusion, fu2024mobile, ze20243d, pertsch2025fast, rouxel2025extremumflowmatching, rouxel2024flowmultisupport}, has enabled robots to learn action policies directly from high-dimensional perceptual observations, such as visual inputs, without requiring an explicit analytical dynamics model. This property is particularly advantageous for robotic tasks involving unstructured environments and complex sensory inputs. Although model-free RL has been applied successfully to soft-object manipulation tasks, including cloth folding \cite{matas2018sim}, purely model-free approaches do not explicitly represent soft-object dynamics or account for latent factors such as material properties and object geometry. This limitation can hinder long-horizon prediction and precise control of object deformation. Conversely, model‑based methods can enable shape control during manipulation through explicit dynamics models, but they require accurate state estimation. Such a requirement is difficult to satisfy under partial observability, and learning such models demands extensive data to cover the large state and action spaces characteristic of soft object manipulation \cite{chen2021ab}.


\subsection{Soft Object World Models}
Modeling soft objects such as garments are especially challenging due to their high-dimensional state space and complex, nonlinear and material-driven dynamics. Recent and prominent research proposes to address the challenges posed by material-driven dynamics by learning soft-object dynamics models directly from empirical data \cite{sanchez2020learning, chen2022comphy, ma2023learning}. The empirical data are often sourced from simulators \cite{mittal2025isaac}, as collecting and curating real world data tends to be labour intensive. However, certain works have instead opted for differentiable simulators, where a small amount of real world data can be used to train for simulating realistic soft object dynamics via gradient optimization \cite{li2022diffcloth}. Such simulators can then be used to generate an even larger corpus of simulated data that more closely match real world dynamics. Another approach is Graph-based neural networks (GNNs), which provide a powerful alternative for modeling the dynamics, demonstrating strong empirical performance across a variety of soft objects such as plasticine \cite{shi2023robocook}, dough \cite{liu2023softgpt}, cloth \cite{lin2022learning, pfaff2020learning}, and fluids \cite{li2018learning, sanchez2020learning}. While these approaches demonstrate impressive results, these methods often require strong domain knowledge and carefully crafted inductive biases, which can restrict their scalability and adaptability to diverse soft objects.

In robotics and vision, world models and predictive state estimators have become central for planning and closed‑loop control, where compact latent representations enable state estimation from partial observations and allow model‑based rollouts for trajectory scoring or MPC \cite{li2025robotic}. However, learning-based dynamics models are required to predict state transitions from interaction data, whereby the choice of state representation is crucial to capture the transitions. Point clouds are selected as the state representation because it balances fine-grained geometric detail and point-wise temporal correspondence with robust sim-to-real transfer from simulators to RGB-D sensors. However, point clouds are inherently high-dimensional, hence they demand a backbone that scales well while preserving fine-grained spatial and temporal consistency, in which Transformers are well suited. Even so, partial observations remain ambiguous as self-occlusion, folds, and viewpoint changes can yield multiple plausible full states, motivating a generative state estimator. Diffusion models \cite{ho2020denoising}, a class of generative models with expressive ability of capturing complex, high-dimensional data distributions, have emerged as a powerful paradigm and has been applied across diverse domains, including generation of images \cite{rombach2022high, peebles2023scalable}, videos \cite{wang2025lavie, ma2024latte}, 3D shapes \cite{poole2022dreamfusion}, as well as robot policy learning \cite{chi2025diffusion, ze20243d} and finally world modeling \cite{alonso2024diffusion, ding2024diffusion, ahmad2026cost}. Motivated by these developments, we investigate the effectiveness of Transformer and generative-based models for predicting soft-object dynamics and supporting manipulation planning.

\begin{figure*}[hbt]
	\centerline{\includegraphics[width=1\linewidth]{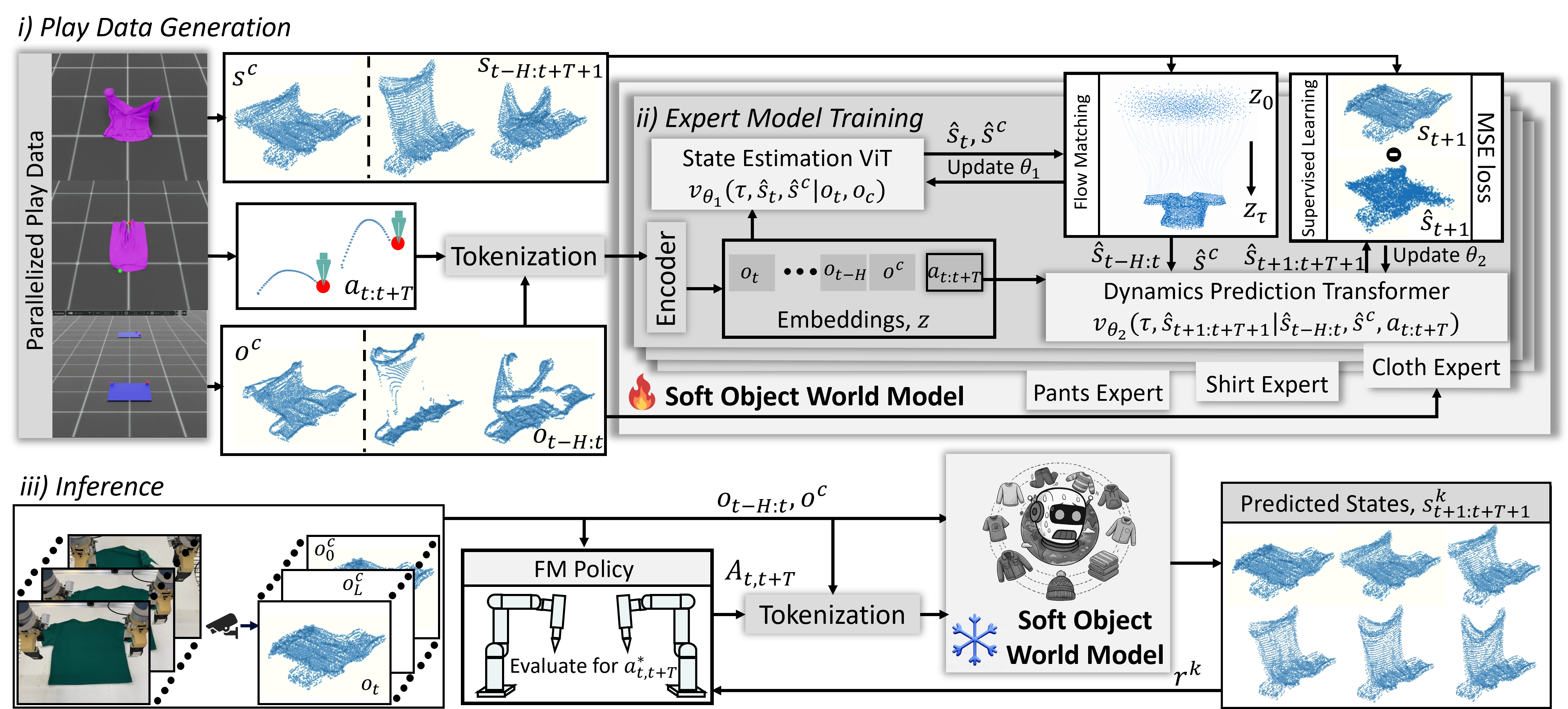}} 
     \vspace{-1.em}
	\caption{Overview of the proposed pipeline, comprising three stages. i) Synthetic soft object manipulation demonstrations are collected in simulation using Isaac Lab, capturing partial point cloud observations and bimanual actions across randomized material properties and grasping configurations. ii) For each garment category, an expert model consists of training two components: a State Estimation Diffusion Model via flow matching; and a Dynamics Prediction Model via supervised learning. iii) Integrate the world model to evaluate action proposals generated via a Goal Conditioned Flow Matching policy.}
	\label{pipeline}
    \vspace{-1.625em}
\end{figure*}

\subsection{In-Context Learning}
In‑context learning (ICL), along with few‑shot test‑time conditioning, has recently emerged as a practical mechanism for adapting models to new instances without weight updates by ingesting short histories or exemplar interactions as context. First introduced in natural language processing with models like GPT‑3 \cite{brown2020language}, ICL has since seen uses in robotics to enable rapid task adaptation with foundation models \cite{di2024keypoint}, identify invariant regions of state space for robust control \cite{zhang2024one}, discover consistent object alignments from few examples \cite{vosylius2023few}, and train policies that generalize across tasks or embodiments \cite{fu2024context,jain2024vid2robot}. Additionally, ICL have recently been used to perform real-time implicit identification for real world dynamics, building upon its potential for dynamics identification \cite{evans2022context}. Building on these ideas, ICL is extended to the soft‑object domain by conditioning a transformer dynamics model on a variable‑length of unrelated estimated states from partial point‑cloud demonstration. This test‑time conditioning lets the world model infer latent material and shape priors from a handful of contextual interactions, and produce materially aware, temporally coherent rollouts that can be used directly for planning and control in soft object manipulation.

\section{Problem Formulation}
We consider a discrete-time robotic manipulation problem in which a bimanual manipulator interacts with a soft object over time steps $t = \{0,1,...,T\}$; the model receives camera depth partial observations history point clouds $o_{t-H:t}$ of  length $H$  and plans manipulation actions $a_{t:t+T}$ at time t, that manipulates the soft object for time length $T$. The learning objective is to train a predictive world model that (i) infers a dense 3-D soft object state $s_t \in \mathbb{R}^{N \times3}$ from partial observations and (ii) generates multi-step, action and context conditional predictions $\widehat{s}_{t+1:t+T+1}$ suitable for planning. To enable the world model to infer the shape and material properties for accurate soft object dynamics prediction, context demonstration $o^c$ of length $L$, will be provided during test time to help the model infer manipulation-relevant properties.

At inference time, the trained dynamics model \(v_{\theta_2}\) receives predicted context \(\widehat{s}^c\),  history observation state $\widehat{s}_{t-H, t}$, a query action trajectory \(a_{t:t+T}\), and a binary grasp mask \(M_t \in \mathbb{R}^{N \times 1}\). Each entry of \(M_t\) indicates whether the corresponding point is grasped on the estimated point clouds, where \(1\) denotes a grasp point and \(0\) denotes a non-grasp point. The predicted future states will then be used to score multiple plausible future action trajectories. All the components of the pipeline benefit from the same large bank of demonstrations.
\vspace{-0.25 em}

\section{Methodology}
The world model as shown in Fig. \ref{pipeline}, has state estimation and dynamics prediction components. Given the camera data of observations history $o_{t-H:t}$ and a context sequence $o_c$, the state estimator infers the full state in the form of a point cloud, producing $\widehat{s}_{t-H:t}$ and $\widehat{s}_c$ for history and context respectively. Conditioned on these estimates and a candidate action plan $a_{t:t+T}$, the dynamics model predicts the ensuing future states $\widehat{s}_{t+1:t+T+1}$. Thus, the world model will need to be trained and conditioned with a large corpus of synthetic demonstrations, and would require a model architecture that can handle high dimensional representations like point clouds. Transformer based models have shown success in predicting full soft object configurations from limited observations. As such, this work will leverage a Transformer for its ability to accept varying-length conditioning and more expressive outputs.

\begin{figure}[t!]
	\centerline{\includegraphics[width=1\linewidth]{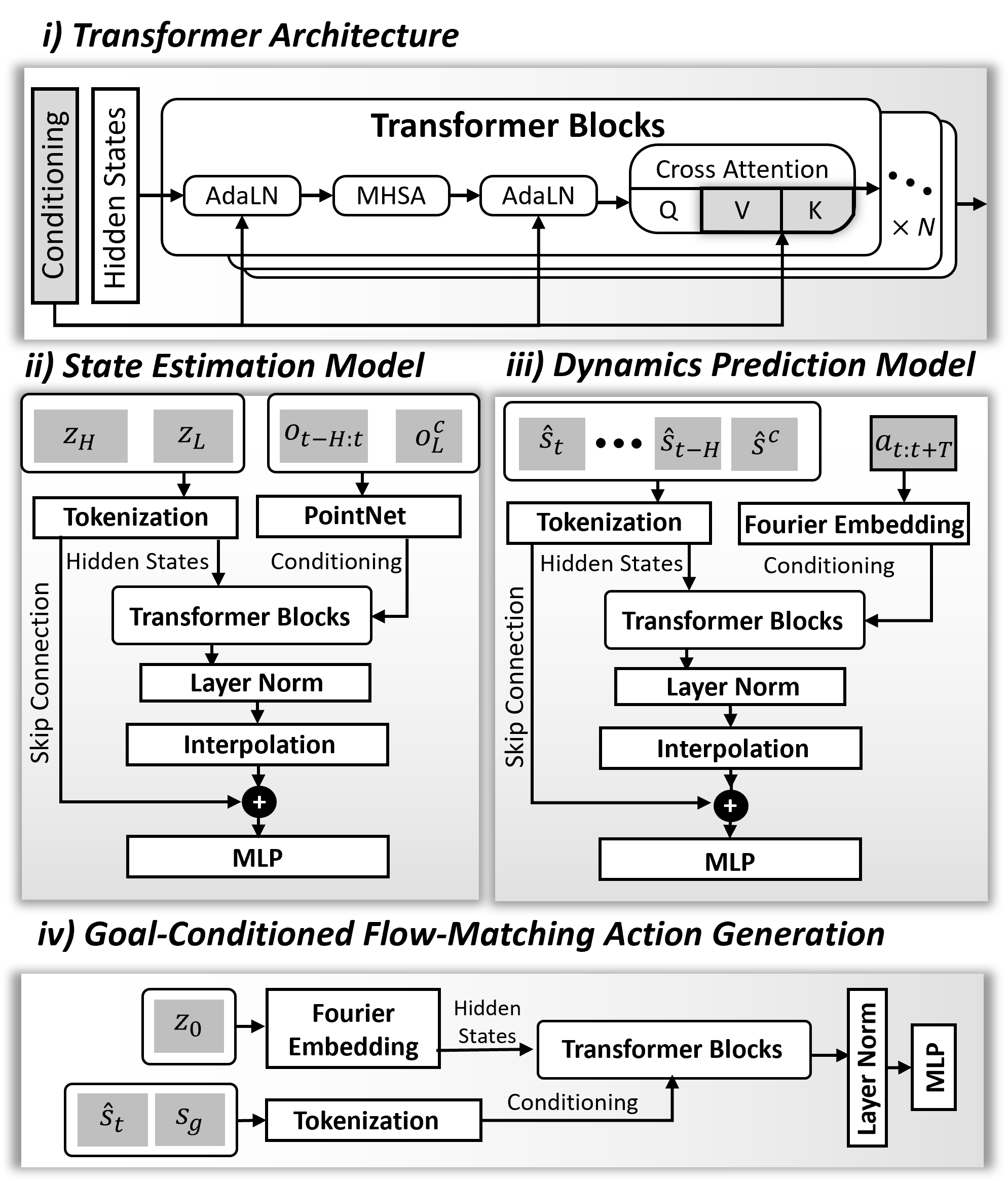}} 
     \vspace{-1.em}
	\caption{Overview of the three Transformer network architectures trained in this work. i) Depicts the transformer architecture which is used in the, ii) state reconstruction model, iii) action-conditioned dynamics prediction model with in-context learning, and the iv) goal-conditioned action-generation model.}
	\label{models_all}
    \vspace{-1.25em}
\end{figure}

\subsection{State Estimation Model}
\label{sec:model}
Within the World Model, the dynamics prediction should be applied on fully observable states which should first be estimated from partial observation. The state estimation model is first formulated as a \emph{conditional flow matching} (CFM) process in which the model is used to reconstruct the corrupted partially observable point cloud $o_{t-H:t}$ and context $o^c$. Conditioning on point clouds reduces the sim-to-real gap because point clouds constitute a mid-level visual representation that preserves geometric invariances across domains. Within the CFM framework, for an arbitrary latent sample $z$, the prediction task is cast as an ordinary differential equation (ODE) of the form:

\begin{equation}
    \label{ODE}
         \begin{small}
    		\begin{aligned}
                \dfrac{d}{d\tau}(z(\tau)) = v(z(\tau),\tau), \quad z(\tau=0) \sim p_0
    		\end{aligned}
    	\end{small}
\end{equation}
where $p_0$ is the arbitrary source distribution, commonly chosen as random Gaussian noise, and $v:\mathbb{R}^d \times [0, 1] \rightarrow \mathbb{R}^d $ is the velocity vector field that induces the probability path from the source to the target distribution. The goal is to learn the vector field that transports the source probability distribution to the data distribution in unit time $\tau$. The vector field $v_{\theta_1}$ can be learned by minimizing the following loss function:
\begin{equation}
    \label{loss}
         \begin{small}
    		\begin{aligned}
                \min_{v_{\theta_1}}\mathbb{E}_{z_0 \sim p_0, z_1 \sim p_1}\left[\int_0^1 {\Vert v_{g\tau}-v_{\theta_1}(z_{\tau}, \tau, o_{t-H:t}, o^c) \Vert}^2_2 d\tau \right]
    		\end{aligned}
    	\end{small}
\end{equation}
where $v_{g\tau}$ is the choice of ground truth vector field to regress against. Here, CFM adopts the time independent, straight line vector field $v_{g\tau} = z_1 - z_0$, which is capable of producing straight probability paths. In practice, the vector field $v_{\theta_1}$ is parametrized by a neural network and the loss in Eq. \ref{loss} can be minimized via gradient descent. In the context of soft object state estimation, the target samples $z_1$ are the desired soft object point cloud states from the expert demonstration dataset. Once the model is trained, $z_0$ is sampled from the source distribution $p_0$ and numerically integrate the ODE from Eq.\ref{ODE} up to time $\tau = 1$ to generate new soft object point cloud states that follow the expert data distribution.

The model architecture backbone used is a vanilla Vision Transformer (ViT) architecture \cite{dosovitskiy2021an} as shown in Fig. \ref{models_all} (i), which demonstrated high scalability in high dimensional state spaces such as image and video generation. For state estimation specifically, the model takes the noisy gaussian source that is transported through the velocity vector field $v_{\theta_1}$ conditioned on the partial observable point cloud. Farthest point sampling (FPS) is then used to sample a fixed number of points, denoted by $P_v$, as patch centers $C \in \mathbb{R}^{P_v\times3} $. To patchify the point cloud, the patch centers $C$ is used to construct a Voronoi diagram in the cartesian space, thus dividing the point cloud into $P_v$ separate regions. Each Voronoi cell is treated as a distinct patch, which will then go through a PointNet \cite{qi2017pointnet} layer for feature extraction and generating the tokens. Following the tokenization process, the input token is directly subjected to a sequence of transformer blocks for processing. For conditioning the point cloud embedding, an adaptive layer normalization (AdaLN) \cite{xu2019understanding} is added to incorporate conditional information, and a cross-attention layer positioned after the multi-head self-attention (MHSA) to incorporate conditional information. In this cross-attention operation, the hidden states serve as the query vector, while the conditional partial point cloud information acts as both the key and value vectors. Lastly, the decoding process transforms the hidden states into 3D point cloud coordinates by first, a distance-weighted interpolation that upsamples the hidden states. Here, the interpolation weights are computed from the distances between points and their corresponding patch centers in the undeformed template soft object, which produces an intermediate representation $x \in \mathbb{R} ^ { P_v\times D}$, where $D$ denotes the feature space \cite{tian2025diffusion}. A Multi-Layer Perceptron (MLP) then maps this representation to the final output $x_{out} \in \mathbb{R} ^ { N\times 3}$, yielding the predicted velocity field at integration time $\tau$. The neural network is then trained to minimize the loss in Eq. \ref{loss}.

\subsection{Dynamics Prediction with In-Context Learning}
The goal of dynamics prediction is to predict the future states of the soft object given robot actions. The state estimation architecture is extended to model dynamics by modifying the conditioning input to incorporate robot actions, as shown in Fig. \ref{models_all} iii). Additional temporal attention layers were added to enhance the temporal modeling \cite{tian2025diffusion}. To learn the conditional posterior distribution $p(\widehat{s}_{t+1:t+T+1}|a_{t: t+T}, \widehat{s}_{t-H:t}, \widehat{s}^c)$, the transformer model was trained via supervised learning. Here, $a_{t: t+T}$ represents the robot actions at time $t$ and prediction horizon of length $T$, $\widehat{s}^c$ denotes the estimated context of length $L$, $\widehat{s}_{t-H:t}$ denotes the estimated state history, and $\widehat{s}_{t+1:t+T+1}$ is the future soft object states to be predicted at timestep $t$, where $H$, $L$  and $T$ can be set at any arbitrary length. The model constructs the future full states by predicting per step deformation field to be added to a last history step, thereby strongly adhereing to the markovian process. A Fourier feature-based embedding following NeRF \cite{mildenhall2021nerf} to represent continuous spatial information is used to effectively encode the action space of $a_{t: t+T} \in \mathbb{R} ^ {T \times 6}$.

However, the dynamics of soft objects such as garments are highly dependent on their geometry, configuration, material properties, and interaction history, all of which can affect how the object deforms under robot actions. In-context learning (ICL) is therefore leveraged to provide test-time interaction data that informs the model of the object and task-specific dynamics required for future-state prediction. The key to implementing ICL is to train the world model on a large corpus of simulated synthetic demonstrations, whereby each demonstration follows a consistent interaction pattern. In this case, the pattern consists of a pick-and-place task for the soft object, where a random pair of pick and place points are sampled from the object and grasped and placed, respectively. Since the model is trained on synthetic data where material parameters are randomized while the interaction pattern is fixed, it learns to associate material-dependent deformations with latent material properties. At inference, the transformer attends to this context and modulates the dynamics rollout without weight updates, enabling effective adaptation to different materials, while supported by improved prediction when the context material matches and degraded performance under mismatched contexts. The world model is trained with a variable length of test time interactions states to emulate the variability in test-time context length. Masking is first used to randomly dropout different lengths of the partial point-cloud conditioning, thereby simulating contexts with varying amounts of available history. The action and partial point-cloud observations are then projected into their respective latent embedding space, and provided to a multi-head self-attention (MHSA) module via cross-attention, along with the Adaptive LayerNorm (AdaLN) layers. This fused representation is then attended by the policy query to produce context-aware predictions of the future soft-object states.

\subsection{Goal-Conditioned Flow-Matching Action Generation}
\label{sec:action_generation}
At inference time, the partial observations ${o}_{t-H:t}$ are mapped to an action trajectory that drives the garment toward the target configuration $s_g$. However, these partial observation point clouds do not contain sufficient information for the policy to infer an optimal trajectory. Instead, they only support the prediction of a statistically meaningful pattern of one, which must be further evaluated by a model to yield accurate action trajectories. Moreover, world models that rely on optimizations that randomly samples actions are inefficient for garment manipulation, as the high-dimensional state space and sensitive deformation dynamics cause small action variations to produce markedly different configurations. Consequently, many feasible candidates are poorly aligned with the current state and goal, making their world-model evaluation costly and uninformative. A goal-conditioned generative policy is therefore used and trained on the same demonstrations, to generate a diverse set of smooth, geometrically meaningful, and task-relevant trajectories. The world model then evaluates these informed proposals and selects the trajectory with the most desirable predicted outcome.

The conditional distribution of actions is modelled as
\begin{equation}
p_{\theta_3}(A_{t:t+T} \mid \hat{s}_{t}, s_g, M_t),
\end{equation}
where $A_{t:t+T}=\{a^0_{t:t+T},\ldots,a_{t:t+T}^K\}$ denotes $K$ number of action trajectory proposals over a planning horizon $T$, and $M_t$ denotes the grasp masks associated with the current garment state. Each action $a_{t:t+T} \in \mathbb{R}^{T \times 6}$ is represented as a relative 6-DoF Cartesian end-effector displacement. The generated trajectory is expressed relative to the grasp points and are transformed into absolute poses by adding them to the corresponding grasp-point poses during execution. This representation ensures that each generated trajectory originates from the selected grasp locations, thereby enforcing physically feasible initial conditions. Consequently, the planner avoids allocating computational resources to evaluating proposals with implausible or poorly aligned initial motions.

\subsubsection{Grasp-Point Selection and Direction Prior}
To identify promising grasp regions, we compute the per-point displacement magnitude between the current and target garment states. Points with larger displacement are more likely to require manipulation and are therefore assigned higher grasp probability:
\begin{equation}
g(\hat{s}_t)=\operatorname{softmax}_{\tau_g}\left(\|\hat{s}_t-s_g\|_2\right),
\end{equation}
where $\tau_g$ is a temperature parameter controlling the concentration of the grasp-point distribution. We sample the grasp points indices $g_k$ with the highest probability from this distribution $g(\hat{s}_t)$. Because the current state and goal state \(s_g\) share consistent point indices, the poses of the selected grasp points can be directly retrieved from both the current state and their corresponding locations in \(s_g\). The displacement between the sampled current and target grasp points defines a goal-directed motion prior:
\begin{equation}
d=(g(s_g)-g(\hat{s}_t)).
\end{equation}

The direction $d$ provides a coarse estimate of the motion required to move the selected garment regions toward their desired positions. This direction is then used to bias the flow-matching gaussian source. Different noise samples produce different initial trajectories, which are transformed by the learned flow into distinct action proposals. This structured initialization guides generation toward the target while preserving diversity across possible manipulation strategies.

\subsubsection{Flow-Matching Policy and World-Model Selection}

We reuse the diffusion-transformer backbone described in Section~IV-A to model action trajectories rather than future garment states. The policy is conditioned on the current state, target state, and grasp masks, which are jointly encoded into a conditioning tensor $S_t$. The model is trained using conditional flow matching over demonstrated action trajectories. Let $z_1=a_{t:t+T}$ denote an expert action trajectory from the dataset. The conditional velocity field $v_{\theta_3}$ is optimized using
\begin{equation}
\mathcal{L}_{\mathrm{act}}=\mathbb{E}_{z_0,z_1,\tau}\left[\left\|(z_1-z_0)-v_{\theta_3}(z_\tau,\tau\mid S_t)\right\|_2^2\right],
\end{equation}
where,
\begin{equation}
z_\tau=(1-\tau)z_0+\tau z_1,\qquad \tau\sim\mathcal{U}(0,1).
\end{equation}
At inference time, the learned flow is integrated from $K$ independently sampled structured sources to generate a diverse set of demonstration-informed action proposals $\{a_{t:t+T}^{(k)}\}_{k=1}^{K}$.

For each proposal, the context-conditioned world model predicts the resulting future garment trajectory:
\begin{equation}
\hat{s}_{t+1:t+T+1}^{(k)}=v_{\theta_2}(\widehat{s}_{t-H:t},\widehat{s}^c,a_{t:t+T}^{(k)}, M_t),
\end{equation}
where $v_{\theta_2}$ denotes the dynamics world model, $\widehat{s}_{t-H:t}$ and $\widehat{s}^c$ are the estimated history and context respectively. The predicted terminal state is compared with the desired target state using the MSE loss:
\begin{equation}
\label{equat}
a_{t:t+T}^*
=
\arg\min_{k\in\{1,\ldots,K\}}
\frac{1}{N}
\sum_{i=1}^{N}
\left\|
\hat{s}_{t+T,i}^{(k)}-s_{g,i}
\right\|_2^2,
\end{equation}

where $N$ is the number of garment points, and $\hat{s}_{t+T,i}^{(k)}$ and $s_{g,i}$ denote the $k$-th predicted and target states, respectively. Only the selected action trajectory $a_{t:t+T}^*$ is executed. By narrowing evaluation to informed proposals, the system reduces unnecessary computation and enables faster action selection.
\begin{figure*}[hbt]
	\centerline{\includegraphics[width=1\linewidth]{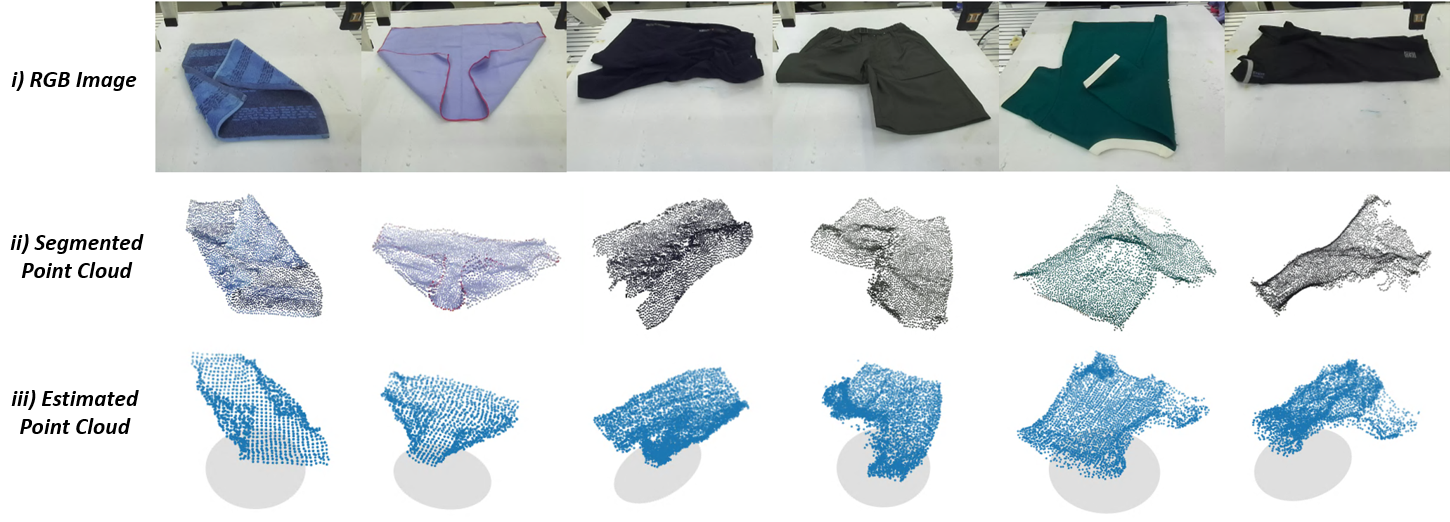}} 
     \vspace{-1.em}
	\caption{ Qualitative results on the state estimation model. i) shows the Raw RGB image from which object segmentation is performed on. ii) shows the segmented point cloud. iii) Shows the estimated point clouds from partial observable segmented point clouds.}
	\label{state_model}
    \vspace{-1.em}
\end{figure*}
\begin{table}[t]
	\centering
	\caption{Quantitative results on state reconstruction}
	\label{state_estimation}
	\begin{threeparttable}
		\begin{tabular}{cccclllll}
			\toprule 
			Category & Method & MSE* & CD* & EMD*\\
			\midrule
                Cloth & MSE Loss & \textbf{0.321} & 4.91 & 5.28\\
                    & UniClothDiff \cite{tian2025diffusion} & 0.682 & 7.76 & 7.18\\
                    & PCN \cite{yuan2018pcn}& 2.08 & \textbf{4.26} & \textbf{4.94}\\
                    & Flow Matching (Ours) & 0.344 & 4.44 & 5.04\\
            \midrule
                Shirt & MSE Loss & 0.696 & 8.08 & 9.72\\
                    & UniClothDiff \cite{tian2025diffusion}  & 0.648 & 5.76 & 6.10\\
                    & PCN  \cite{yuan2018pcn} & 2.48 & \textbf{1.56}& \textbf{2.42}\\
                    & Flow Matching  (Ours) &\textbf{0.150} & 1.91 & 2.95\\
            \midrule
                Average & MSE Loss & 0.509 & 6.50 & 7.50\\
                    & UniClothDiff \cite{tian2025diffusion}  & 0.665 & 6.76 & 6.64\\
                    & PCN \cite{yuan2018pcn}& 2.28 & \textbf{2.91}& \textbf{3.68}\\
                    & Flow Matching  (Ours) &\textbf{0.247} & 3.18 & 4.00\\
		    \bottomrule[1pt]
		\end{tabular}
            	\begin{tablenotes}[para, flushleft]
			\item[*] The loss are all scaled in $10^-2$. 
		\end{tablenotes}
	\end{threeparttable}
\vspace{-1.5em}
\end{table}

\section{Experiment}
\subsection{Setup}We evaluate our method in both simulation and real-world environments. Specifically, we use Isaac Lab as the simulation platform for data collection and evaluation, and demonstrate effective sim-to-real transfer in the real-world setting. 

\textbf{Simulation:} The simulation is built upon Isaac Lab, leveraging its GPU-accelerated parallel environment capability to simulate large volumes of soft object manipulation trajectories efficiently. Soft object dynamics are simulated using the built-in PhysX-based solver, with soft object deformation handled by the projective dynamics solver. Training data are generated by randomizing garment material properties, including bending, stretching and damping coefficients, as well as trajectory and bimanual grasping configurations, to maximize the coverage of the state and action spaces and mitigate the sim-to-real gap. The training set comprised 100,000 samples of randomized folding demonstrations for each soft object, each with 110 timesteps. Soft-object states and action trajectories are obtained from simulation, while partial point clouds are segmented from simulated camera observations. Each demonstration was synthesized by selecting two pairs of pick and place points at random and planning an arc trajectory between each pick–place pair to produce two bimanual folding stages. 

\textbf{Real World Experiments:} The real-world evaluation is conducted across two robot platforms to demonstrate cross-embodiment generalization of the proposed pipeline. The first platform is X-trainer, a 6+6-DoF bimanual robot arm system operable via a master-arm teleoperation interface for context demonstration collection. The second platform is the Tiangong humanoid robot, representing a full-body anthropomorphic embodiment with 42 degrees of freedom. Additionally, both platforms utilizes different grippers, which are a 2 DOF pincher and a soft hand respectively, thus introducing different grasping dynamics. The inclusion of these two morphologically distinct platforms serves to validate that the proposed pipeline is agnostic to robot embodiment, as the action representation is parameterized as relative 6-DoF end-effector displacements and the world model operates exclusively on garment point cloud states, requiring no platform-specific architectural modifications or retraining.

\textbf{Evaluation Metric:} 
To compare the predicted states of the real garment and simulated during identical manipulation paths, the mean squared error (MSE), Chamfer Distance (CD) and Earth Movers Distance (EMD) between the observed points and the corresponding simulated particles are calculated from the same manipulation trajectory. MSE measures point-wise correspondence error, which is critical for the downstream dynamics prediction, while CD and EMD evaluates the global geometric similarity such as shape.

\begin{figure*}[hbt]
	\centerline{\includegraphics[width=1\linewidth]{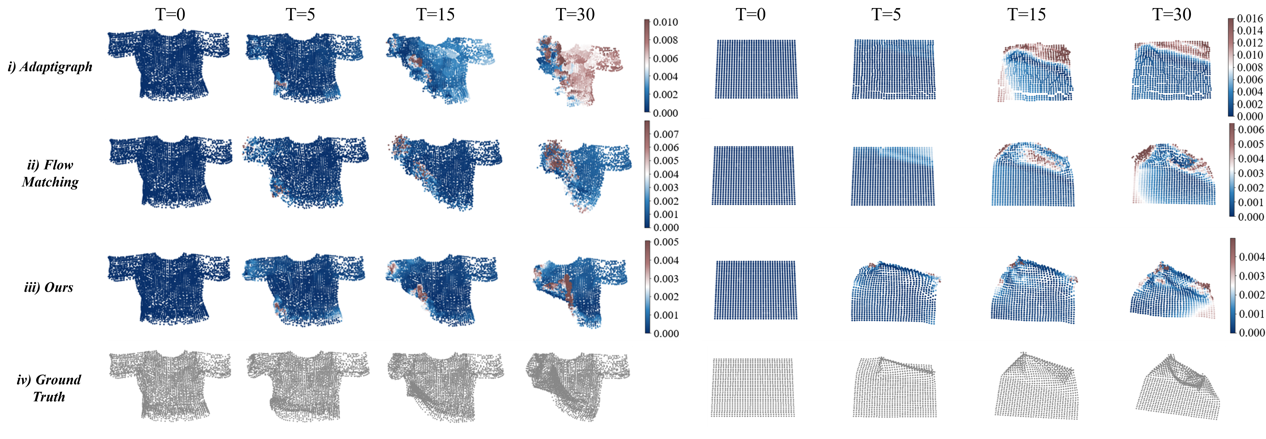}} 
     \vspace{-1.em}
	\caption{Qualitative comparison of dynamics predictions of the shirt (left) and cloth (right) produced by the i) GNN, ii) Generative flow matching, and (iii) supervised Transformer model against the (iv) ground-truth states. The colorbar indicate the MSE loss between the baseline predictions and ground truth.}
	\label{dyn_comp}
    \vspace{-1.5em}
\end{figure*}
\begin{table}[t]
	\centering
	\caption{Quantitative results on Dynamics Prediction}
	\label{dynamics_comp}
	\begin{threeparttable}
		\begin{tabular}{ccccclllll}
			\toprule 
			Category & Method & MSE* &CD*& EMD*\\
			\midrule
                Cloth & MSE Loss (Ours) & \textbf{0.137} & 4.97 & 4.70\\
                    & UniClothDiff \cite{tian2025diffusion} & 0.185 & 5.00 & 4.77\\
                    & Flow Matching & 0.180 & 4.89 & 4.97\\
                    & Adaptigraph \cite{zhang2024adaptigraph} & 0.157 & \textbf{4.83} & \textbf{4.31}\\
            \midrule
                Shirt & MSE Loss (Ours)& \textbf{0.347} & \textbf{7.07} & \textbf{7.72}\\
                    & UniClothDiff \cite{tian2025diffusion}& 0.463 & 7.88 & 8.36\\
                    & Flow Matching  & 0.531 & 8.28 & 8.68\\
                    & Adaptigraph \cite{zhang2024adaptigraph} & 0.415 & 8.03 & 8.69\\
            \bottomrule[1pt]
		\end{tabular}
            	\begin{tablenotes}[para, flushleft]
			\item[*] The loss are all scaled in $10^-2$. 
		\end{tablenotes}
	\end{threeparttable}
\vspace{-1.5em}
\end{table}

\subsection{State Reconstruction}
Training the state estimation model required approximately eight hours per epoch on an GTX 3090 GPU. Ground-truth and partial point clouds were sampled randomly from the generated dataset. Fig. \ref{state_model} presents state estimation results, including the RGB input, segmented partial point clouds, and predictions from the ViT-based state estimation model. The baselines considered in the comparative study include a Point Completion Network (PCN) \cite{yuan2018pcn}, and Transformer models trained via UniCloth's Diffusion \cite{tian2025diffusion}, Flow Matching (FM) and MSE loss. Qualitatively, the estimated shapes from partial observations closely resemble the corresponding real-world soft objects, suggesting that the transformer can recover global object geometry from occluded visual observations.

Comparison studies are conducted in simulation and reported in Table \ref{state_estimation}, with each method evaluated over 100 trials. PCN achieve the lowest CD and EMD loss across the evaluated settings. However, PCN produces much higher mean squared error (MSE), where its MSE for the shirt category is approximately 20 times greater than that of the FM model. MSE is particularly important for the subsequent dynamics prediction stage because point-wise temporal correspondence must be preserved. Therefore, FM provides the best overall performance, achieving the lowest MSE while remaining highly competitive with PCN in terms of CD and EMD.

\subsection{Dynamics Prediction}
Training the dynamics prediction model required approximately one day per epoch. The dataset was identical to that used for state estimation, except that each batch contained temporally consecutive point clouds sampled from the same demonstration sequence. Additionally, the material parameters of every demonstration are recorded during data generation. At each iteration, a target demonstration is sampled together with an additional context demonstration whose material parameters match those of the target, and the model is trained to predict the target's future state sequence conditioned on this independent context. To reduce memory and compute requirements while preserving the prediction horizon, all simulation data were subsampled with a fixed stride, specifically at uniformly spaced intervals of 5 frames. The baselines considered in the comparative study include Adaptigraph's GNN-based model \cite{zhang2024adaptigraph}, and Transformer models trained via UniCloth's Diffusion \cite{tian2025diffusion}, Flow Matching (FM) and MSE loss. 

Comparison studies are conducted in simulation and reported in Table \ref{dynamics_comp}, with each method evaluated over 100 trials. The Transformer model trained on MSE supervised loss achieved the lowest loss across both object categories. Although its Chamfer Distance (CD) and Earth Mover's Distance (EMD) were higher than those of the GNN model for the cloth category, it achieved lower CD and EMD for the shirt category. This suggests that the GNN has greater difficulty modelling objects with increased geometric complexity, as reflected by its higher CD and EMD relative to the other generative methods in the shirt category. Qualitative results in Figure \ref{dyn_comp} further show that the MSE-trained Transformer and Generative Flow Matching models generate more physically plausible point clouds. Overall, the MSE-trained Transformer better captures the underlying dynamics than generative objectives as the simulated dynamics are deterministic, fully observable, and Markovian, and therefore do not exhibit multimodality. 

\begin{figure}[t!]
	\centerline{\includegraphics[width=1\linewidth]{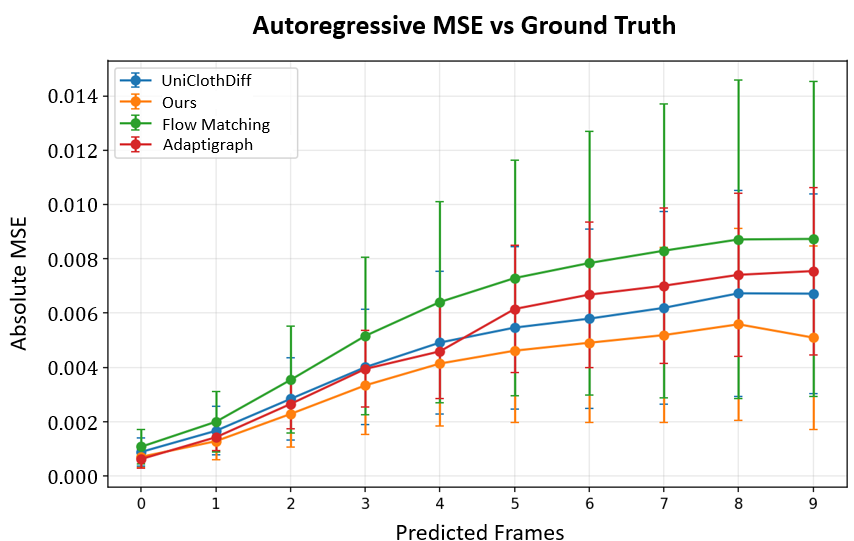}} 
     \vspace{-1.em}
	\caption{Autoregressive prediction errors of Diffusion, Flow Matching, GNN, and classical Transformer models, measured using mean squared error (MSE) against the ground-truth states.}
	\label{autoregressive_loss}
    \vspace{-1.em}
\end{figure}

Figure \ref{autoregressive_loss} shows that the autoregressive performance of all models degrades as the prediction horizon increases, with the GNN showing the fastest degradation. On average, the MSE trained Transformer and GNN model achieved stronger performance, likely due to the deterministic nature of their supervised learning formulations. This is consistent with dynamics prediction, in which a fixed initial condition and fully observable state should yield a single unique future trajectory.

\begin{figure}[t!]
	\centerline{\includegraphics[width=1\linewidth]{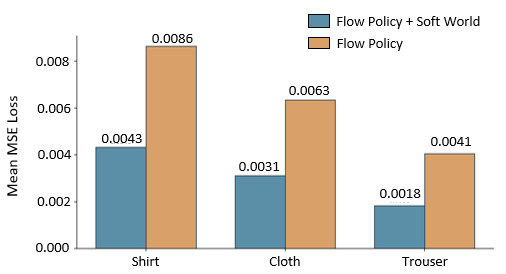}} 
     \vspace{-1.em}
	\caption{Ablation study comparing the mean simulated terminal-state MSE of trajectories from the Goal-Conditioned FM Policy with and without the World Model Evaluator against desired state, across all garment categories.}
	\label{policy_ablation}
    \vspace{-.75em}
\end{figure}
\begin{figure}[t!]
	\centerline{\includegraphics[width=1\linewidth]{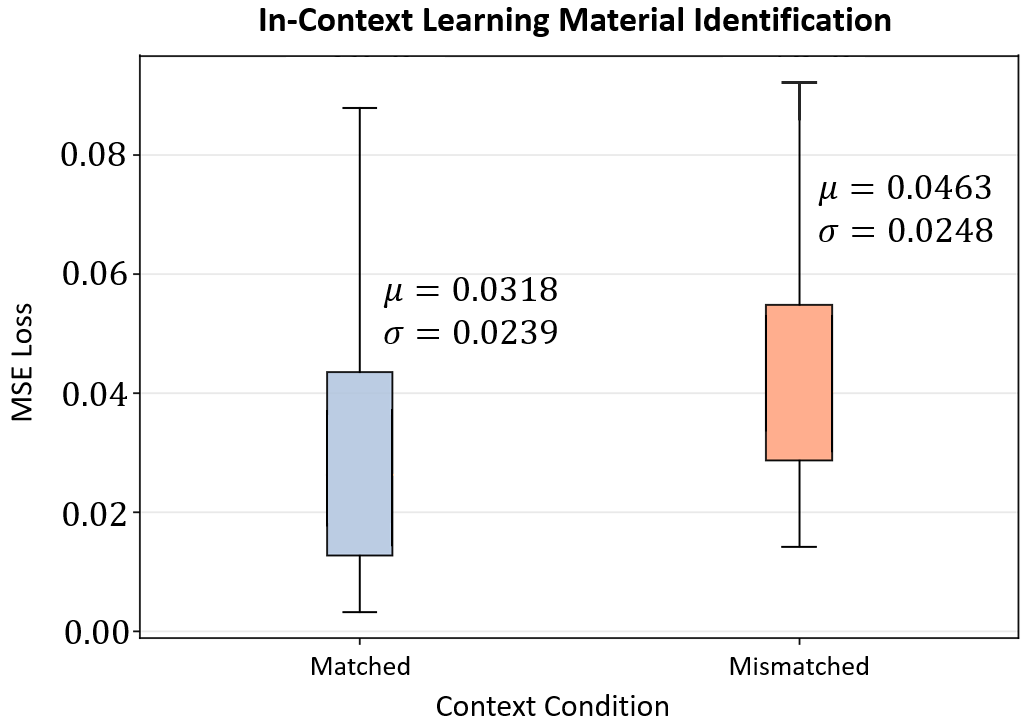}} 
     \vspace{-.5em}
	\caption{Ablation study comparing the MSE loss of predictions generated using matched and scrambled material parameters in the context conditioning.}
	\label{ablation_mh}
    \vspace{-1.25em}
\end{figure}



\begin{figure*}[t!]
	\centerline{\includegraphics[width=1\linewidth]{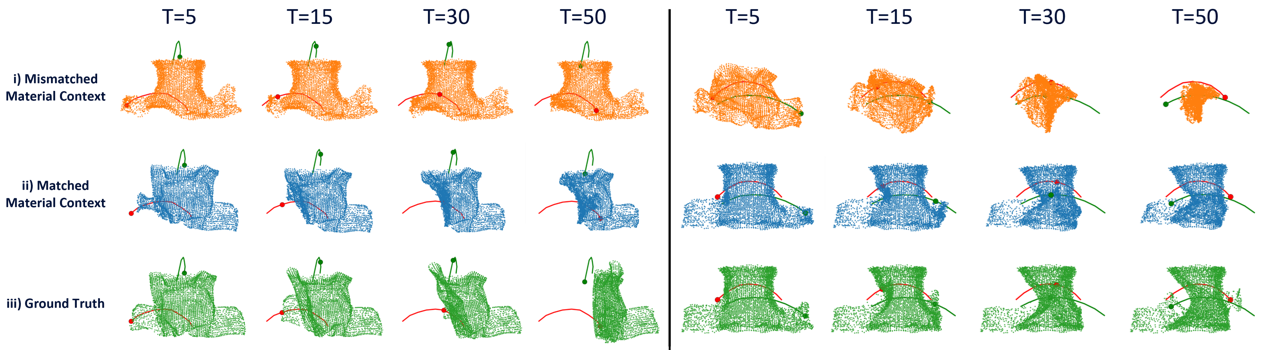}} 
     \vspace{-.5em}
	\caption{ Visualization of the effect of material context on predicted object dynamics. Predicted soft-object states when conditioned on (i) mismatched and (ii) matched material parameters, compared against (iii) ground truth. The left and right blocks correspond to two interaction variants: the left uses material parameters more resistant to bending and stretching, while the right uses material parameters that bend and stretch more readily.}
	\label{mh_visual}
    \vspace{-1.em}
\end{figure*}

\subsection{Ablation Studies}
This section investigates whether the world model provides a performance boost as an action proposal evaluator. For each garment category, 100 simulated trials are run in which the Goal-Conditioned FM policy generates a set of candidate action trajectories toward a desired goal state. Two conditions are compared: the policy executing a candidate proposal without evaluation, versus where the world model rolls out each proposal and selects the one predicted to reach the target most closely. The planned trajectories are simulated in both conditions, and the terminal timestep is compared against the desired state using MSE. As shown in Fig. \ref{policy_ablation}, the policy equipped with the World Model Evaluator achieves a lower terminal-state error across all garment categories, roughly 2x the performance relative to the unevaluated policy. This shows that the world model performs a form of reasoning over the action proposals, ranking them by their predicted outcomes rather than accepting a single sampled trajectory, and thereby selecting proposals that drive the garment closer to the target configuration. A single real world trial is also conducted and shown in Fig. \ref{dyn_comp}(i)(ii), showing that the world model evaluated action proposal is indeed closer to the target state.

\begin{figure*}
	\centerline{\includegraphics[width=1\linewidth]{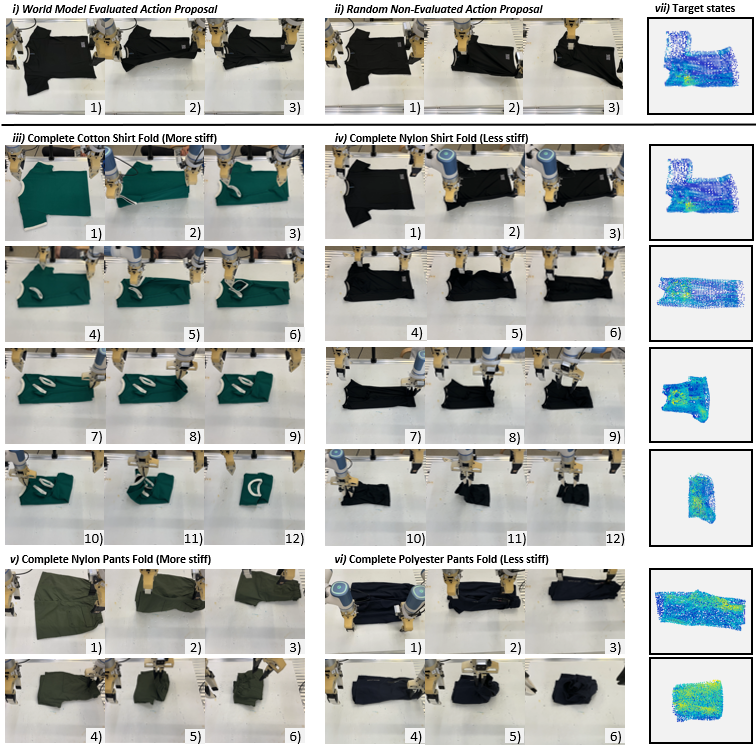}} 
     \vspace{-1.em}
	\caption{ Qualitative results on real world full garment folds.  Part (i) shows the performance of the goal-conditioned policy with the world model, while part (ii) shows its performance without the world model. Parts (iii-vi) are the four stage full fold for the (iii)(1-12) green cotton shirt, (iv)(13-24) black nylon shirt. And the two stage full fold for the (v)(1-6) green nylon trousers. (iv)(7-12) black polyester trousers. Part (vii) shows the target states for each manipulation.}
	\label{full_fold}
    \vspace{-1.5em}
\end{figure*}

\subsection{In-Context Learning Material Identification}
This section investigates whether in-context dynamics identification from a short demonstration sequence of excited states is sufficient to infer soft object material properties. For this study, the dynamics model is evaluated under two conditions. In the first condition, the material parameters associated with the context demonstration are consistent with those of the ground truth but with different grasp points and manipulation. In the second condition, the material parameters of the context demonstration and ground truth are randomly mismatched. Prediction accuracy is quantified using MSE loss with respect to ground truth. As shown in Fig. \ref{ablation_mh}, the proposed in-context method exhibits a 1.5x performance increase between the matched and mismatched material-parameter conditions. This result suggests that in the considered folding scenarios, the context provides sufficient information for distinguishing material properties. Additionally, qualitative results in Fig. \ref{mh_visual} show a clear visual improvement in the predicted object states when using matched material context.



\subsection{Real World Rollouts}

Real-world experiments were conducted across three garment categories: flat cloth, shirts, and trousers. For each category, two garments of different material variants and sizes were evaluated. Ten trials were performed for each garment configuration, resulting in a total of 60 real-world rollouts. The trials include desired states that require different manipulation coordination modes, such as bimanual coordinated, bimanual sequential and single arm manipulation. Before each trial, \(L=4\) teleoperated demonstrations were collected to provide the ICL conditioning input, \(o^c\) and history \(o_{t-H:t}\). Each demonstration consists of partial point-cloud observations acquired using a static RGB-D camera and segmented into partially observable point clouds. No gradient-based parameter updates were applied during evaluation. The evaluated objects varied in both geometry and material properties.

During execution, the State Estimation Model reconstructs the full test  time interaction observations, \(\hat{s}_{t-H:t}\), from the partially observed point-cloud demonstration, \(o_{t-H:t}\). Conditioned on \(\hat{s}_{t-H:t}\), the Dynamics Model generates forward rollouts for the candidate action trajectories \(\{A_{t,t+T}^{(k)}\}_{k=1}^{K}\) produced by the goal-conditioned flow-matching policy. The candidate trajectory that minimizes the objective defined in Eq.~\ref{equat} is selected for execution. After execution, the final partial point cloud is segmented from the RGB-D observation and compared with the corresponding simulated target state using the Chamfer distance (CD) and Earth Mover's Distance (EMD). A manipulation trial is considered successful when the resulting loss is below a predefined threshold. The success rates over all 60 real-world trials are reported in Table~\ref{real2sim_results}.

\subsection{Real World Long Horizon Folding}
To ensure the method's ability to perform long-horizon garment folding, all models were trained using demonstrations involving two-fold sequences. During long horizon folding evaluation, the models were used to predict and assess trajectories comprising between two and four sequential folds. Shirts were subjected to four-fold sequences, thereby requiring the model to generalize to trajectories containing two additional folds beyond those observed during training. Trousers were evaluated using two-fold sequences and yielded satisfactory performance. Qualitative results are presented in Fig.~\ref{full_fold}.

\begin{figure*}
	\centerline{\includegraphics[width=1\linewidth]{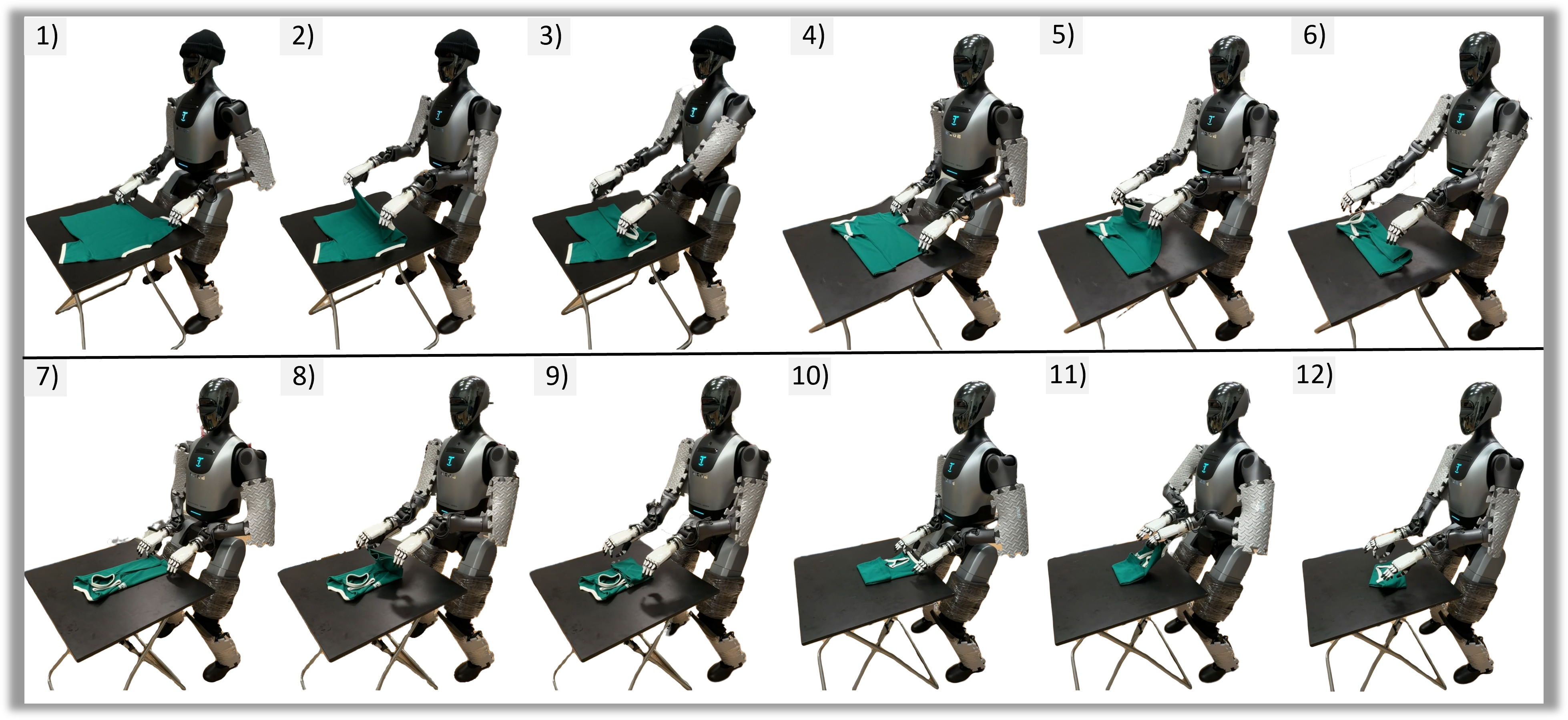}} 
     \vspace{-1.em}
	\caption{Qualitative results from real-world experiments in which a humanoid robot folds a cotton shirt using soft hands.}
	\label{tk}
    \vspace{-1.em}
\end{figure*}

Additionally, by leveraging an embodiment-agnostic action space design, the method enables effective cross-embodiment transfer. The Fig. \ref{tk} showcases a successful real world cross embodiment from the 7 DOF bimanual robot arm with parallel grippers to a 42 DOF humanoid robot with dexterous hands.

\begin{table}[t]
	\centering
	\caption{Quantitative results on Real World Rollouts}
	\label{real2sim_results}
    \vspace{-1.em}
	\begin{threeparttable}
		\begin{tabular}{cccc}
			\toprule 
            Category & Success Rate & CD ($10^-2$)& EMD ($10^-2$)\\
			\midrule
                Square Cloth & 9/10 & 2.54 & 2.81\\
                Rectangular Cloth & 9/10 & 2.34 & 3.10\\
                Nylon Trouser  & 8/10 & 2.96 & 4.09\\
                Polyester Trouser  & 10/10 & 2.71 & 3.60\\
                Cotton Shirt  & 10/10 & 2.92 & 4.29\\
                Nylon Shirt  & 9/10 & 2.83 & 4.24\\
		    \bottomrule[1pt]
		\end{tabular}
	\end{threeparttable}
\vspace{-1.5em}
\end{table}

\section{Discussion}

The results show that proposed method reconstructs partially observed soft object point clouds, predicts action-conditioned deformations, and enables real-world folding across diverse garments. The MSE-Trained Transformer achieved the lowest MSE and slower autoregressive error growth than the generative and GNN baselines. This suggests that deterministic supervised prediction is well suited to modeling deterministic, fully observable, and Markovian dynamics, where a fixed estimated state, observation context, and action sequence generally correspond to a single unique future prediction. Generative models may instead be more suitable for multimodal tasks, including state estimation under occlusion and action proposal generation. Although PCN achieves favorable CD and EMD, flow matching attains lower MSE, which is important for preserving the point-wise correspondence required for deformation prediction and goal evaluation. Additionally, the ICL results support its benefit, where the predicted states show a clear difference in material properties and behavior when conditioned on different material contexts. Real-world results shows high success across cloths, shirts, trousers, materials, and two robot embodiments without test-time updates, supporting sim-to-real transfer. This is enabled by point clouds representation, simulation randomization, and relative Cartesian actions. Successful multi-stage folding beyond the two-fold training demonstrations further indicates generalization to longer task sequences, although errors can accumulate.

 Future work should introduce increasingly dynamic motions, including rapid pulling, shaking, lifting, release, and flinging, to further investigate the capability of the context conditioning under tasks with more dynamic deformations. It should also test whether deterministic supervised dynamics models remain preferable as future-state ambiguity increases through uncertain contacts or partial observations. Finally, incorporating repeated observation, reconstruction, and replanning in a closed-loop framework could mitigate accumulated prediction and execution errors in long-horizon manipulation.

\section{Conclusion}
This paper presents an In-Context Learning Soft Object World Model framework for bimanual garment manipulation under partial observations. This work combines generative point-cloud state reconstruction, in-context action-conditioned dynamics prediction, and goal-conditioned motion proposal generation. Together, these components estimate garment geometry, model the dynamics of soft objects, to evaluate and select among candidate action proposals for real world bimanual garment folding. Experiments in simulation showed that the supervised Transformer provides accurate and stable dynamics predictions while accurately inferring material behaviour from context conditioning, and that the reconstruction module produces correspondence-preserving full-state estimates suitable for downstream prediction and planning. Real-world experiments across cloths, shirts, trousers, material variants, and two robot embodiments demonstrated successful sim-to-real transfer and effective generalization, while achieving high task success without additional real-world training. These results suggest that In Context Learning World Models are a practical approach for generalizable garment manipulation. The current study focuses primarily on quasi-static folding, where material-dependent effects may be less pronounced than geometric and configuration-dependent cues. Future work will investigate more dynamic garment interactions, evaluate the role of material-history correspondence under stronger dynamic excitation, and examine deterministic and generative dynamics models under increasingly ambiguous future states. Incorporating closed-loop state estimation and replanning will also be important for reducing accumulated errors during longer-horizon manipulation tasks.

\bibliographystyle{ieeetr}
\bibliography{ref_rp}

\end{document}